# A Survey on Deep Learning of Small Sample in Biomedical Image Analysis

Pengyi Zhang[1,2], Yulin Deng[1,2], Xiaoying Tang[1,2], Yunxin Zhong[1,2], Xiaoqiong Li[1,2*]

[1]School of Life Science, Beijing Institute of Technology, Haidian District, Beijing, P.R. China

[2]Key Laboratory of Convergence Medical Engineering System and Healthcare Technology, Ministry of Industry and Information Technology, Haidian District, Beijing, P.R. China

**Abstract**

The success of deep learning has been witnessed as a promising technique for computer-aided biomedical image analysis, due to end-to-end learning framework and availability of large-scale labelled samples. However, in many cases of biomedical image analysis, deep learning techniques suffer from the small sample learning (SSL) dilemma caused mainly by lack of annotations. To be more practical for biomedical image analysis, in this paper we survey the key SSL techniques that help relieve the suffering of deep learning by combining with the development of related techniques in computer vision applications. In order to accelerate the clinical usage of biomedical image analysis based on deep learning techniques, we intentionally expand this survey to include the explanation methods for deep models that are important to clinical decision making. We survey the key SSL techniques by dividing them into five categories: (1) explanation techniques, (2) weakly supervised learning techniques, (3) transfer learning techniques, (4) active learning techniques, and (5) miscellaneous techniques involving data augmentation, domain knowledge, traditional shallow methods and attention mechanism. These key techniques are expected to effectively support the application of deep learning in clinical biomedical image analysis, and furtherly improve the analysis performance, especially when large-scale annotated samples are not available. We bulid demos at https://github.com/PengyiZhang/MIADeepSSL.

## 1. Introduction

Biomedical imaging techniques have been widely used in clinical practice. Specialists used to manually analyze biomedical images and further fit all clinical clues together to make the final diagnosis based on their own experience. Nowadays, the manual biomedical image analysis is facing four major challenges: (1) Manual analysis is limited by individual experience and thus diagnosis might be different among individuals. (2) Training a qualified specialist requires lots of money and years of efforts. (3) Rapid growth of biomedical images in both number and modality puts enormous pressure on specialists. (4) Repetitive and tedious analysis work on unpleasing biomedical images makes specialists feel tired easily, which might lead to a delayed diagnosis or even a misdiagnosis and thus pose a great risk to patients. These challenges exacerbate shortage of healthcare resources to a certain extent especially in under-developed regions. An alternative solution is computer-aided biomedical image analysis.

Driven by the growth of computing power and availability of large-scale labelled samples (e.g., ImageNet(Deng et al., 2009) and COCO(Lin et al., 2014)), deep neural network has been extensively studied due to its fast, scalable and end-to-end learning framework. Especially, Convolution Neural Network (CNN)(LeCun et al., 2015) models have achieved significant improvements compared with traditional shallow methods in image classification (e.g., ResNet(He et al., 2016) and DenseNet(G. Huang et al., 2017)), object detection (e.g., Faster R-CNN(Ren et al., 2017) and SSD(Liu et al., 2016)) and semantic segmentation (e.g., UNet(Ronneberger et al., 2015) and Mask R-CNN(He et al., 2017)), etc. As biomedical imaging equipment is increasingly applied and popularized in clinical practice, large numbers of biomedical images have been accumulated. Therefore, deep learning techniques have been gradually applied in computer-aided biomedical image analysis, such as diabetic retinopathy detection(Gulshan et al., 2016), skin cancer diagnosis(Esteva et al., 2017), pulmonary disease detection based on X-ray radiography(Rajpurkar et al., 2018) and heart disease risk prediction in fundus retinal images(Poplin et al., 2018), etc. Computer-aided biomedical image analysis using deep learning techniques has been one of the most active research topics. Some studies(Esteva et al., 2017; Rajpurkar et al., 2018) claim that their biomedical image analysis applications based on deep learning techniques have achieved expert-level or even beyond expert-level performance. However, in many cases of biomedical image analysis, deep learning techniques suffer from small sample learning (SSL) dilemma caused mainly by lack of annotations. Labeling multimodal biomedical images is a highly specialized work that experienced biomedical experts are required to perform fine-grained annotations, which puts extra enormous pressure on biomedical experts. This is a major reason why biomedical images usually lack adequate annotations in comparison of large-scale natural images. With inadequate training samples, deep models are prone to overfitting, thus leading to a big drop in performance. Therefore, how to apply deep learning techniques effectively on a small biomedical image set becomes an urgent problem to be well studied.

---

*Corresponding author

Many surveys(Litjens et al., 2017; Ker et al., 2018; Anwar et al., 2018; Razzak et al., 2018; Esteva et al., 2019; Maier et al., 2019; Moen et al., 2019) review the progress of deep learning techniques applied in biomedical image analysis from four directions: (1) deep models; (2) analysis tasks, e.g., classification, detection and segmentation; (3) modalities of biomedical images; (4) anatomical structure, e.g., brain, chest and retina. This SSL dilemma is simply mentioned as a challenge and its solution is not surveyed comprehensively. Different from these studies, our survey aims to deal with the SSL dilemma that deep learning techniques generally suffer from in biomedical image analysis tasks. Shu et al.(2018) investigated SSL methods in big data era by dividing them into concept learning and experience learning. Their study theoretically clarified the rationality of the SSL regime by providing neuroscience evidences and furtherly presented a general view of SSL techniques for various application scenarios. To be more practical, in this paper we survey key SSL techniques for deep learning of small sample in biomedical image analysis by combining with the development of relevant techniques in computer vision applications. In order to support the clinical usage of biomedical image analysis based on deep learning techniques, we intentionally expand our survey to include explanation methods for deep models that are important to clinical decision making.

In this paper, we divide our surveyed key SSL techniques into five categories (Fig. 1):

(1) **Explanation techniques**. The clinical applications require that the decision-making process of deep models is explainable and transparent. We thus explore the potential procedure of computer-aided diagnosis system applied in clinical practice and survey the explanation techniques for deep models in **Section 2**.

(2) **Weakly supervised learning techniques**. Since the fine-grained annotations required by fine-grained biomedical image analysis tasks (e.g., lesion localization and gland segmentation) are costly, can we just use relatively low-cost and coarse-grained annotations to perform these fine-grained tasks in biomedical image analysis? Therefore, the weakly supervised learning techniques for dealing with such problem are discussed in **Section 3**.

(3) **Transfer learning techniques**. Deep learning has made significant progress in solving various domain related problems. Can the knowledge gained by deep models while solving problems in other domains be applied in biomedical image analysis and thus alleviate the demands for large-scale annotated biomedical images? We present transfer learning techniques for answering this question in **Section 4**.

(4) **Active learning techniques**. Not every annotated sample contributes equally to deep models. Can we just select the few most valuable samples for annotations and train deep models with them for biomedical image analysis? We introduce active learning techniques for solving this problem in **Section 5**.

(5) **Miscellaneous techniques**. Miscellaneous techniques that help improve the performance of deep models in biomedical image analysis are studied in **Section 6**, involving data augmentation, domain knowledge, shallow methods and attention mechanism.

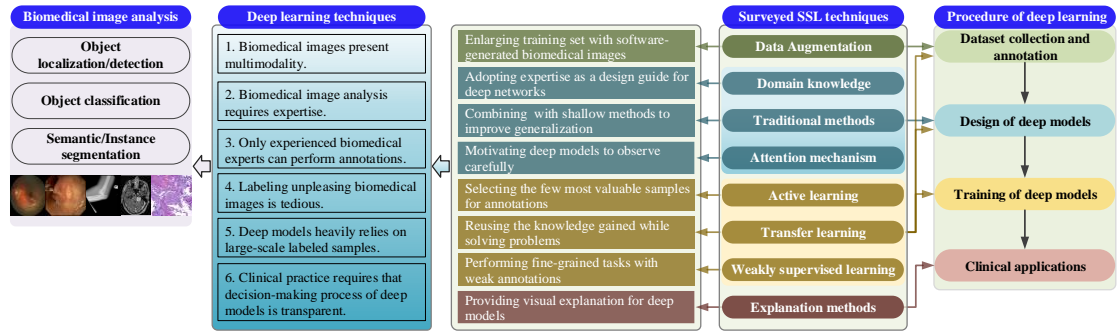

Figure 1. Overview of our surveyed SSL techniques for deep learning in clinical biomedical image analysis.

To our best knowledge, this is the first attempt to present a comprehensive survey on key SSL techniques that are expected to effectively support the application of deep learning in clinical biomedical image analysis and furtherly improve analysis performance especially in small sample cases. Notably, our survey focusing on key SSL techniques can be mutually complementary with previous reviews on deep learning in biomedical image analysis.

## 2. Explanation techniques

Due to highly abstract feature representation and end-to-end learning framework, decision-making process of deep models remains largely unclear(Petsiuk et al., 2018). In clinical practice, an incorrect decision might do great harm to patient's physical and mental health. Explaining decision-making process of deep models can provide decision-making evidences for doctors' final diagnosis to enroll doctors in the decision-making process of computer-aided biomedical image analysis (Fig. 2a) and make the clinical decision more reliable. Thus, clinical applications require that the decision-making process of deep models is explainable and transparent, which is vital to build a trusty computer-aided diagnosis system. Existing explanation methods for deep models can be divided into local or global explanation and white-box or black-box methods (Fig. 2b). Table 1 outlines some representative studies on explanation methods.

**CAM-based methods**. One of the well-known explanation methods is Class Activation Map (CAM)(Zhou et al., 2016). This method constructs CAMs by calculating the class-specific activation in each spatial location of the last convolutional feature maps and tries to clarify the classification decision-making process by providing visual explanation that highlights class-specific and discriminative regions. In particular, Rajpurkar et al. (2018)(2017) adopted this approach to explain deep models for abnormality detection in musculoskeletal X-rays (Fig. 3a) and pulmonary disease classification in chest X-rays. Compared with the captions provided by a board-certified radiologist, visual CAMs presented a high relevance to clinical abnormality. Essentially, CAM takes fully use of the last convolutional features and the class-specific weights of output layer, thus making it an efficient method for visual explanations. Other similar methods adopted different pooling methods (e.g., global max pooling, log-sum-exp pooling, and log-mean-exp pooling(B. Zhang et al., 2018)) to retrieve different visual CAMs. Besides, Selvaraju et al.(2017)proposed Grad-CAM by using gradients instead of weights of output layer to calculate CAMs. Grad-CAM greatly expands the applications of CAM without any modifications to deep models and without restriction to classification models.

**Table 1**. Representative researches on explanation methods for deep models.

| Reference | Category | Method & remark |
|---|---|---|
| Erhan et al. (2009) | Global explanation, White-box | Activation maximization; gradient ascent in the input space |
| Simonyan et al. (2013) | Global explanation, White-box | Activation maximization; gradient ascent in the input space |
| | Local explanation, Black-box | Examine component importance; perturbing input image pixels |
| Zeiler et al. (2013) | Local explanation, White-box | Examine component importance; covering up different portions of input image with a gray square |
| Bach et al. (2015) | Local explanation, White-box | Backpropagation-based; class-specific error |
| Yosinski et al. (2015) | Global explanation, White-box | Activation maximization; a regularized optimization in image space |
| Zhou et al. (2016) | Local explanation, White-box | CAM-based; multiplying activation with the weights of output layer |
| Mahendran et al. (2016) | Local explanation, White-box | Backpropagation-based; visualizing features by mapping them back to the input pixel space |
| Zhang et al. (2016) | Local explanation, White-box | Backpropagation-based; class-specific error |
| Ribeiro et al. (2016) | Local explanation, Black-box | Examine component importance; perturbing the pixels of input image randomly to learn a linear explainable decision model |
| Selvaraju et al. (2017) | Local explanation, White-box | CAM-based; extending CAM by replacing the weights of output layer with the gradients |
| Fong et al. (2017) | Local explanation, Black-box | Examine component importance; learning meaningful perturbation on input image by gradient decent |
| Nguyen et al. (2017) | Global explanation, White-box | Activation maximization; a optimization regularized by a conditional generative adversarial network |
| Esteva et al. (2017) | Local explanation, White-box | Backpropagation-based; L1 norm of loss gradients to input layer |
| Rajpurkar et al. (2018) | Local explanation, White-box | CAM-based; explaining a deep model for abnormality detection in musculoskeletal X-rays |
| Rajpurkar et al. (2018) | Local explanation, White-box | CAM-based; explaining a deep model for abnormality detection in chest X-rays. |
| Zhang et al. (2018) | Local explanation, White-box | CAM-based; modifing CAM with different global pooling methods |
| Kermany et al. (2018) | Local explanation, White-box | Examine component importance; convolving an occluding kernel across the input image to find the clinically relevant regions of age-related macular degeneration and diabetic macular edema |
| Petsiuk et al. (2018) | Local explanation, Black-box | Examine component importance; perturbing input image with random masks to calculate the average mask weighted by class-specific output |

**Backpropagation-based methods**. The other type of explanation methods is built on backpropagation of class-specific gradients(Zeiler and Fergus, 2014; Mahendran and Vedaldi, 2016) or errors(Bach et al., 2015; Jianming Zhang et al., 2018). The visual explanation from backpropagation-based method is generally sharper than that from CAM-based methods. In particular, Esteva et al.(2017) adopted such a backpropagation-based approach to explain a deep model for skin cancer classification in dermoscopy images. The generated saliency maps demonstrated a highly clinical relevance to the skin cancer diagnosis (Fig. 3b).

**Examining component importance**. Another type of explanation methods(Petsiuk et al., 2018; Erhan et al., 2009; Simonyan et al., 2013; Ribeiro et al., 2016; Fong and Vedaldi, 2017; Kermany et al., 2018) examine the importance of each component for decision-making by removing or modifying the components (e.g., pixel, image patch and segment) in an input image. The visual explanation is retrieved by combining each component weighted by its importance together. In particular, Kermany et al.(2018) developed such an explanation method to explain deep models for the classification tasks of age-related macular degeneration and diabetic macular edema. The generated occlusion maps highlighted pathological areas in retina OCT images, presenting a highly clinical relevance (Fig. 3c).

**Activation maximization**. For global explanation, activation maximization(Erhan et al., 2009) is generally performed as an optimization problem by gradient ascent algorithm, where the parameters required to be updated are image pixels instead of the weights of deep models. For better visualization, some studies(Yosinski et al., 2015; Nguyen et al., 2017) have tried to add various regularization constraints on activation maximization to make the optimization results more realistic.

Existing explanation methods for deep models are expected to help build trusty computer-aided diagnosis systems and accelerate such diagnosis systems to be implemented in clinical practice. The explanation results, i.e., visual attention map or saliency map, can also be used for weakly supervised object localization and instance segmentation to alleviate the demands for fine-grained annotations. However, a key limitation across these studies is that they cannot explain how robust a decision is, e.g., when deep models will fail in a specific biomedical image analysis task, which is very important to construct automatic and unattended diagnosis system. Besides, the majority of these methods are designed for CNN-based recognition tasks. When there exists multiple targets in one image, the produced visual explanation is prone to highlighting only the few most discriminative targets. The incompleteness of visual explanation might lead to a sub-optimal result. Future work likely focus on developing explanation methods for deep models in complex tasks to explain how robust the decisions are by digging deep into clinical expertise underlying biomedical images. Moreover, the cooperative mode how biomedical image analysis using deep learning techniques effectively help clinical specialists is also a research topic worthy of exploring.

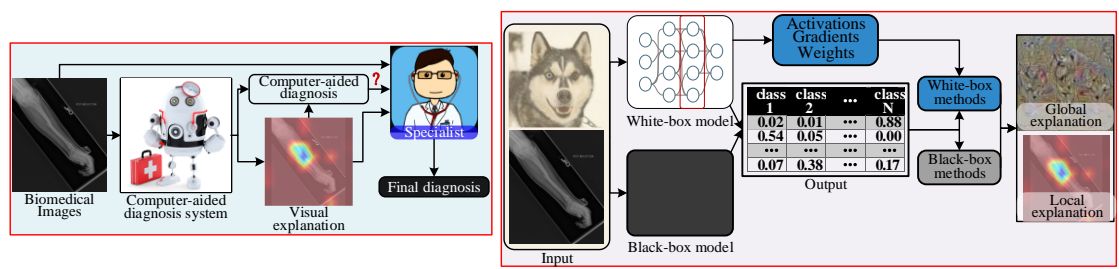

Figure 2. Left-to-right: (a) A potential procedure of computer-aided diagnosis system applied in clinical practice. Explanation methods for the decision-making process are expected to help build trusty computer-aided diagnosis systems. (b) Taxonomy of explanation methods. For instance, the global explanation methods aims to explain a model by visualizing the model's favorite "husky", while the local explanation methods try to explain why a model think a input image is a "husky" by highlighting the regions that the model relies on to make the decision. The black-box models build the explanation with only the input and output of a model while the white-box explanation methods construct the explanation by using the input and output as well as internal information, such as activations, gradients and weights, etc.

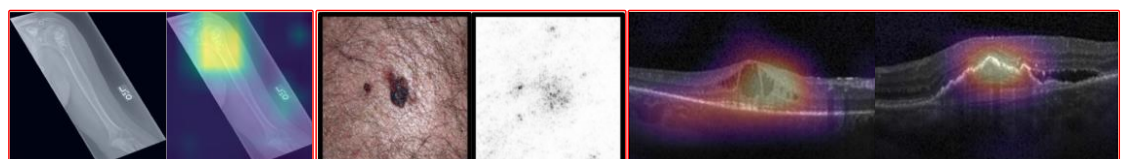

Figure 3. Examples of visual explanation for deep models built on three different modalities of biomedical images. Left-to-right: (a) Visual CAMs. The visual CAMs(Rajpurkar et al., 2017) highlight the abnormal region in musculoskeletal X-rays, presenting a highly clinical relevance; (b) Saliency maps. In skin cancer classification based on dermoscopy images, the saliency maps(Esteva et al., 2017) describe the importance of each pixel for diagnosis, presenting a highly clinical relevance; (c) Occlusion maps. The occlusion maps(Kermany et al., 2018) highlight the clinically relevant areas of pathology for the classification task of age-related macular degeneration and diabetic macular edema in retina OCT images.

## 3. Weakly supervised learning techniques

With the popularity of high resolution biomedical images in clinical diagnosis, many fine-grained tasks have been introduced in biomedical image analysis, such as lesion localization and gland segmentation. Training deep models for these fine-grained tasks generally requires fine-grained annotations that are costly and difficult to acquire. A natural idea is to use relatively low-cost and coarse-grained annotations to perform these fine-grained tasks in biomedical image analysis. Weakly supervised learning is such a promising approach to alleviate the demands for fine-grained annotations.

Weakly supervised learning(Zhou, 2018) mainly addresses learning problems for incomplete, inaccurate and inexact supervision environments. In computer vision, the research frontier problems of weakly supervised deep learning have started a gradual transition from simple object localization tasks(Zhou et al., 2016; Fong and Vedaldi, 2017) to complex object detection tasks(Bilen and Vedaldi, 2016; Tang et al., 2017; Diba et al., 2017; Tang et al., 2018; Wan et al., 2018; Arun et al., 2019) and semantic segmentation tasks(Z. Huang et al., 2018; Li et al., 2018; Kervadec et al., 2019). The supervision modalities in weakly supervised learning also present a diversified development trend, involving image-level supervision(Bilen and Vedaldi, 2016), box supervision(Bonechi et al., 2018), scribble supervision(Lin et al., 2016; Can et al., 2018), point supervision(Bearman et al., 2016; Kozuka and Niebles, 2017), and webly supervision(Navarro et al., 2018; Jin et al., 2017), etc. The performance of weakly supervised learning methods in various computer vision tasks has been significantly improved, revealing a tendency to catch up with the fully supervised learning algorithms. Taking object detection task as an example, the mean average precision (mAP) score of weakly supervised object detection has been increasingly improved on the PascalVOC 2007(Everingham et al., 2010) dataset (Fig. 4). The gap of mAP (i.e., mean of average precision) scores between weakly supervised object detection algorithms and a well-known fully supervised object detection algorithm, i.e., YOLO(Redmon et al., 2016), is less than 10%. Therefore, the weakly supervised learning techniques are expected to achieve excellent performance in fine-grained biomedical image analysis using weak annotations.

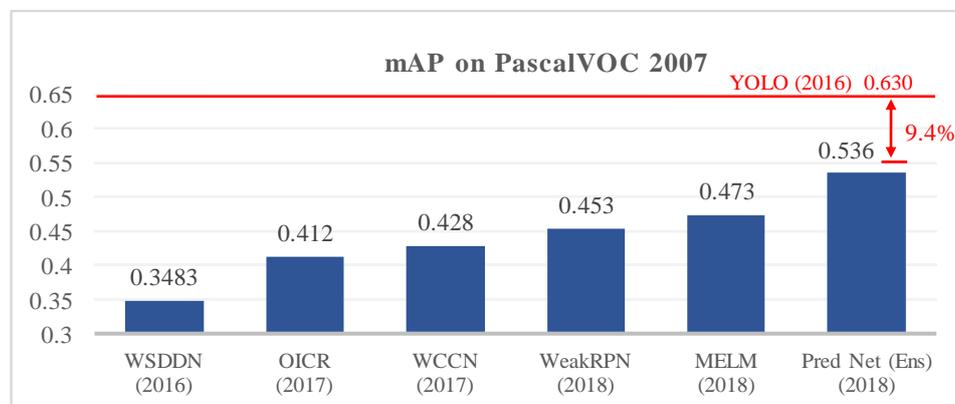

Figure 4. Illustration of the progress of weakly supervised object detection methods, including WSDDN(Bilen and Vedaldi, 2016), OICR(Tang et al., 2017), WCCN(Diba et al., 2017), WeakRPN(Tang et al., 2018), MELM(Wan et al., 2018) and Pred Net (Ens)(Arun et al., 2019).

**Table 2.** Representative papers for weakly supervised deep learning techniques used in biomedical image analysis.

| Reference | Supervision modality | Method | Task | Dataset |
|---|---|---|---|---|
| Gondal et al. (2017) | Image-level supervision | CAM-based attention map + thresholding method | Localization of Diabetic Retinopathy (DR) lesions | 88,702 fundus images in Kaggle Dataset (https://www.kaggle.com/c/diabetic-retinopathy-detection) and 89 fundus images in DiaretDB1 Dataset(Kauppi et al., 2007) |
| Quellec et al. (2017) | Image-level supervision | Backpropagation-based attention map + thresholding method | Localization of Diabetic Retinopathy (DR) lesions | 88,702 fundus images in Kaggle Dataset and 89 fundus images in DiaretDB1 Dataset |
| Rajchl et al. (2017) | Box supervision | CNN-based classifier + dense conditional random field (DenseCRF), iteratively updating the learning target classes for input patches | Brain and lung segmentation | MR images of 55 fetal subjects(Damodaram et al., 2012) |
| Can et al. (2018) | Scribble supervision | Region growing on scribble annotations + UNet + CRF, iteratively refining UNet | Cardiac segmentation and prostate segmentation | 200 volumes in cardiac ACDC(Bernard et al., 2018) and 29 volumes in prostate NCI-ISBI(Bloch et al., 2015) |
| Fernando et al. (2018) | Webly-supervision | Augmenting dermoscopy images from the Internet and using transfer learning to deal with the natural noise of webly-supervision | Skin lesion classification | 1300 skin lesion images(Ballerini et al., 2013) |
| Yang et al. (2018) | Box supervision | Combining fully convolutional network (FCN) for coarse segmentation and graph search (GS) for fine segmentation | Gland segmentation | 14 whole-slide clinical H&E stained histology images of human intestinal tissues(Zhang et al., 2016a) |
| | | | Lymph node segmentation | 207 clinical ultrasound images of human neck lymph nodes(Zhang et al., 2016b) |
| | | | Fungus segmentation | 84 images captured by serial block-face scanning electron microscopy(Zhang et al., 2017) |
| Qu et al. (2019) | Point supervision | Combining Voronoi diagram and k-means clustering methods for coarse segmentation and modified UNet based on DenseCRF loss for fine segmentation | Nuclei Segmentation | 40 images from 8 different lung cancer cases in Lung Cancer dataset and 30 image including a diversity of nuclear appearances from seven organs in MultiOrgan dataset(Kumar et al., 2017) |
| Kervadec et al. (2019) | Point supervision | ENet based on inequality (lower and upper bounds on size) constrained loss function | Cardiac segmentation | 100 cine magnetic resonance (MR) exams covering: dilated cardiomyopathy, hypertrophic cardiomyopathy, myocardial infarction with altered left ventricular ejection fraction and abnormal right ventricle (https://www.creatis.insa-lyon.fr/Challenge/acdc/) |
| | | | Vertebral body segmentation | 23 3D T2-weighted turbo spin echo MR images from 23 patients (http://dx.doi.org/10.5281/zenodo.22304) |
| | | UNet based on inequality (lower and upper bounds on size) constrained loss function | Prostate segmentation | Prostate transversal T2-weighted MR images of 50 patients (https://promise12.grand-challenge.org) |

Many studies have started to apply weakly supervised learning techniques to train deep models for biomedical image analysis tasks (Table 2). These applications mainly include weakly supervised object localization(Gondal et al., 2017; Quellec et al., 2017) and weakly supervised semantic segmentation(Kervadec et al., 2019; Can et al., 2018; Rajchl et al., 2017; Yang et al., 2018; Qu et al., 2019) , where the supervision modalities involve image-level supervision(Gondal et al., 2017; Quellec et al., 2017), scribble supervision(Can et al., 2018), box supervision(Rajchl et al., 2017; Yang et al., 2018), and point supervision(Qu et al., 2019; Kervadec et al., 2019). Lesion localization is very important to clinical diagnosis. For instance, accurate localization of Diabetic Retinopathy (DR) lesions in fundus retina photos is of great help to improve the performance of DR lesion recognition. To implement weakly supervised localization, researchers(Gondal et al., 2017; Quellec et al., 2017) first train a CNN-based classifier with image-level annotations (e.g., whether a fundus photo contains DR lesions or not). Second, they retrieve attention maps or saliency maps that highlight the discriminative and lesion-related regions in biomedical images through the explanation methods for this CNN-based classifier. Finally, the lesion areas are localized by performing thresholding methods on attention maps or saliency maps (Fig. 5a). For weakly supervised segmentation tasks, scribble, box and point supervision modalities are generally used in biomedical image analysis. A representative solution is an iterative optimization with three steps (Fig. 5b), including (1) pre-processing for initial pseudo pixel-level labels (e.g., region growing(Can et al., 2018), Voronoi diagram and k-means clustering methods(Qu et al., 2019)), (2) training deep segmentation model (e.g., UNet(Kervadec et al., 2019; Can et al., 2018; Qu et al., 2019), FCN(Yang et al., 2018) and ENet(Kervadec et al., 2019)) with pseudo pixel-level labels, and (3) post-processing (e.g., GS(Yang et al., 2018) and DenseCRF(Can et al., 2018; Rajchl et al., 2017; Qu et al., 2019)) for refined pseudo pixel-level labels.

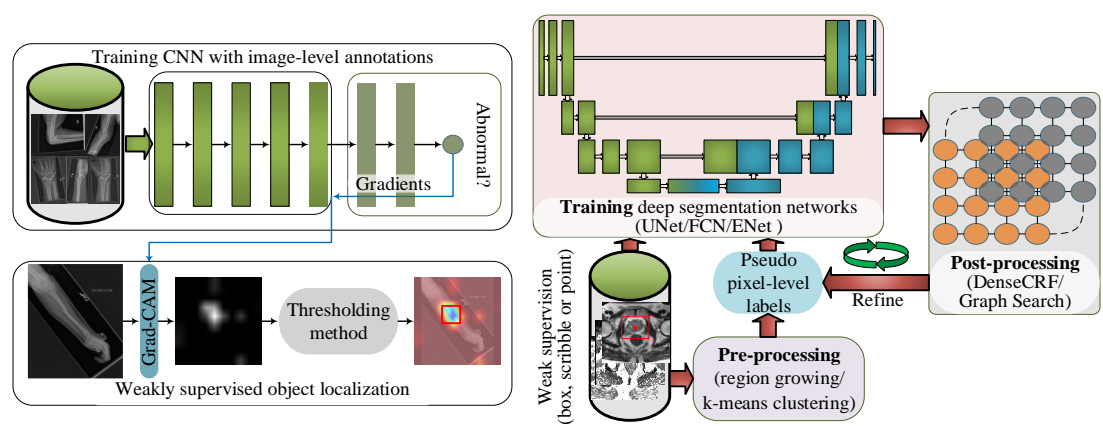

Figure 5. Illustration of weakly supervised learning techniques. Left-to-right: (a) A representative solution for weakly supervised object localization using image-level annotations. (b) A representative for weakly supervised semantic segmentation using box, scribble or point supervision.

Many successful applications have demonstrated that weakly supervised learning techniques are expected to relieve the urgent demands of deep models for fine-grained annotations in many fine-grained biomedical image analysis tasks. Although the performance of weakly supervised learning has been notably improved, there still exists a gap between weakly supervised learning and fully supervised learning that requires to be furtherly studied. Besides, how to utilize the results of weakly supervised tasks to improve the performance of original tasks might be a potential and extended research topic.

## 4. Transfer learning techniques

Labeled biomedical images are scarce, while large-scale natural image sets with various levels of annotations are publicly available, such as ImageNet, COCO, and PascalVOC, etc. A natural idea is whether the annotated samples from other domains can help improve deep models in biomedical image analysis. Unfortunately, when the deep models trained in source domain (e.g., natural images and synthetic images) are applied to the target domain, i.e., biomedical images, the performance of deep models generally drops a lot. Transfer learning is a key technique to solve the problem of model adaptation across domains.

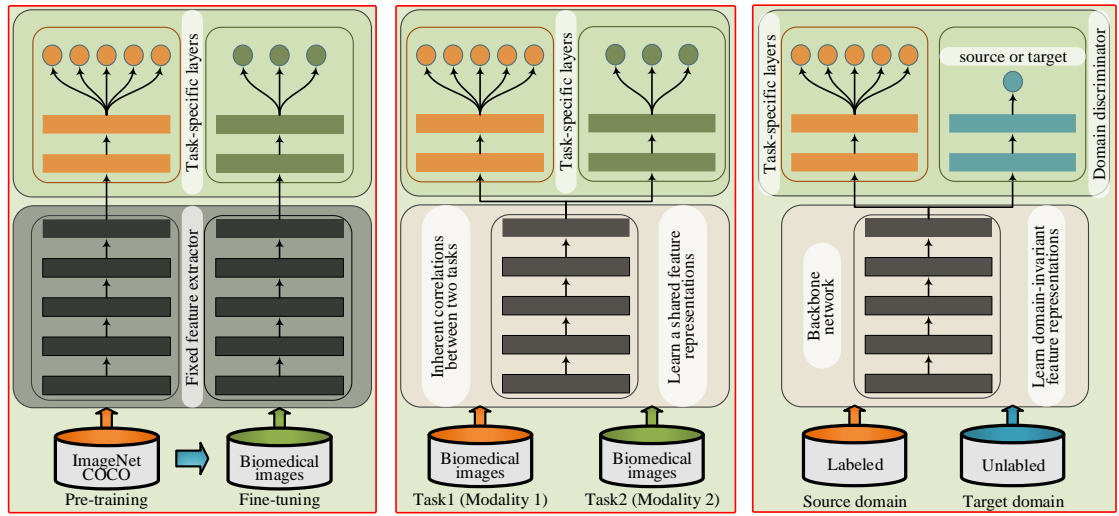

Figure 6. Illustration of transfer learning techniques. Left-to-right: (a) Fine-tuning. Deep models are initially pre-trained on ImageNet or COCO and sequentially fine-tuned on biomedical images by fixing some layers and updating weights of task-specific layers at a small learning rate. (b) Multi-task learning. Deep models for multi-task learning that share some layers and differ in task-specific layers are trained alternatively across different but related tasks. (c) Adversarial training. A domain discriminator branch network is added to standard deep model to motivate backbone network to learn domain invariant feature representation. The entire model is optimized alternatively similarly with multi-task learning.

Building every deep model from scratch is time-consuming and relies heavily on large-scale annotated samples. Transfer learning techniques focuses on extracting the knowledge gained while solving problems in source domain and applying it to different but related problems in target domain. For convenience and efficiency, the image sets with large-scale annotated samples are prone to being selected as source domain in many computer vision applications. To sum up, the transfer learning methods have roughly experienced five stages development(Zhang, 2019), including instance re-weighting adaptation(Huang et al., 2007), classifier adaptation(Lixin Duan et al., 2012), feature adaptation(Xu et al., 2016), deep network adaptation(Long et al., 2015) and adversarial adaptation(Chen et al., 2018). The research frontier problems of transfer learning techniques have been gradually moved from simple image classification tasks(Lixin Duan et al., 2012) to complex and fine-grained tasks, such as object detection(Chen et al., 2018; Inoue et al., 2018) and semantic segmentation(Yang Zhang et al., 2017; Kamnitsas et al., 2017a), etc. Compared with deep models trained from scratch in target domain or trained in source domain but applied directly in target domain, deep models using transfer learning techniques generally achieve much better performance in various computer vision tasks. Therefore, transfer learning might be a promising technique for applying deep learning methods effectively to biomedical image analysis tasks, especially in the small sample cases.

Many studies have adopted transfer learning techniques to improve deep models in biomedical image analysis tasks (Table 3), mainly involving fine-tuning(Kermany et al., 2018; Tajbakhsh et al., 2016; Zhou et al., 2017; Zhou et al., 2019), multi-task learning(Moeskops et al., 2016; Liao and Luo, 2017; Samala et al., 2017) and adversarial training(Dong et al., 2018; Moeskops et al., 2017; Javanmardi and Tasdizen, 2018).

Fine-tuning might be one of the most popular transfer learning techniques. This method (Fig. 6a) generally adapts deep models (e.g., AlexNet(Tajbakhsh et al., 2016; Zhou et al., 2017; Zhou et al., 2019) and InceptionV3(Kermany et al., 2018)) pre-trained on ImageNet to biomedical image analysis by freezing some layers in pre-trained deep models while retraining the others (task-specific layers). The fixed layers are supposed to encode the knowledge gained while solving problems in source domain and act as a fixed feature extractor. During fine-tuning process, the weights of deep models are generally updated at a small learning rate nearby the values of pre-trained weights, equivalent to narrowing the search space of parameters. Thus, fine-tuning method is expected to reduce the dependency of deep models on large-scale annotated biomedical images. Existing studies(Kermany et al., 2018; Tajbakhsh et al., 2016; Zhou et al., 2017; Zhou et al., 2019) consistently reveal that deep models trained by fine-tuning method are more robust and accurate than counterparts trained from scratch in various biomedical image analysis tasks.

Multi-task learning aims to exploit the inherent correlations amongst different but related tasks, such as similar tasks in different biomedical imaging modalities(Moeskops et al., 2016; Samala et al., 2017), and different tasks in same biomedical imaging modalities(Liao and Luo, 2017). The deep models modified for multiple tasks (e.g., ImageNet DCNN(Samala et al., 2017) and ResNet-50(Liao and Luo, 2017)) generally share some layers to learn a same feature representation but differ in task-specific layers to perform multiple tasks (Fig. 6b). Thus, these multi-task deep models can be jointly trained on multiple image sets for multiple biomedical image analysis tasks. In comparison of respective single-task learning, multi-task learning techniques in these studies(Moeskops et al., 2016; Liao and Luo, 2017; Samala et al., 2017) are prone to achieving better results. It implies that multi-task learning techniques have great potential to effectively apply train deep models in biomedical image analysis when training samples from a single modality are limited.

Table 3. Representative papers using transfer learning techniques in biomedical image analysis based on deep models.

| Reference | Transfer learning techniques | Tasks | Dataset | Deep models |
|---|---|---|---|---|
| Tajbakhsh et al. (2016) | Fine-tuning | Polyp detection | 40 short colonoscopy videos | AlexNet |
| | | Pulmonary embolism detection | 121 CT pulmonary angiography datasets with a total of 326 PEs | |
| | | Colonoscopy frame classification | 6 complete colonoscopy videos | |
| | | Intima-media boundary segmentation | 92 carotid intima-media thickness videos | |
| Moeskops et al. (2016) | Multi-task learning | Anatomical structure segmentation | 34 MR brain images, 34 MR breast images and 10 cardiac CT angiography scans | Customized CNN model |
| Zhou et al. (2017) | Fine-tuning | Colonoscopy frame classification | 6 complete colonoscopy videos | AlexNet |
| | | Polyp detection | 38 short colonoscopy videos | |
| | | Pulmonary embolism detection | 121 CT pulmonary angiography datasets with a total of 326 PEs | |
| Moeskops et al. (2017) | Adversarial training | Brain MRI segmentation | 35 MR brain images from adult subjects and 20 axial MR brain images from elderly subjects | Customized CNN model |
| Samala et al. (2017) | Multi-task learning | Classification of malignant and benign breast masses | 1,655 SFM views and 310 DM views with 2454 masses (1057 malignant, 1397 benign) | ImageNet DCNN |
| Liao et al. (2017) | Multi-task learning | Skin lesion classification | 21657 dermoscopy images that contain both the skin lesion and body location labels | ResNet-50 |
| Zhou et al. (2018) | Fine-tuning | Carotid intima-media thickness video interpretation | 92 carotid intima-media thickness videos from 23 patients | AlexNet |
| Kermany et al. (2018) | Fine-tuning | Age-related macular degeneration and diabetic macular edema classification | 207,130 retina OCT images | Inception V3 architecture |
| | | Pneumonia Detection | 5,232 chest X-ray images from a total of 5,856 patients | |
| Dong et al. (2018) | Adversarial training | Chest organ segmentation | 247 grayscale chest X-rays and 221 grayscale chest X-rays | FCN-based model |
| Mehran et al. (2018) | Adversarial training | Eye vasculture segmentation | 40 eye fundus images | FCN-based model |
| | | Neuron membrane detection | 100 electron microscopy images of size 1024×1024 and 125 electron microscopy images of size 1250×1250 | |

Domain adversarial training techniques try to enforce deep models to learn a domain invariant representation (e.g., deep features) for related tasks across different domains (e.g., two biomedical image sets from different sources(Dong et al., 2018; Javanmardi and Tasdizen, 2018)) or motivate deep models to generate outputs close to ground-truth(Dong et al., 2018; Moeskops et al., 2017). In general, these techniques are implemented by adding a branch network of domain discriminator to a standard deep model, where domain discriminator and standard deep model share one backbone network (Fig. 6c). Thus, similar to multi-task learning, the entire model can be optimized alternatively across different domains. Domain adversarial training techniques are expected to reduce the distribution discrepancy of deep features(Dong et al., 2018; Javanmardi and Tasdizen, 2018) (or outputs(Dong et al., 2018; Moeskops et al., 2017)) across domains and thus improve the performance of deep models in biomedical image analysis even without the requirement of annotated samples in target domain.

This key technique for deep learning of small sample in biomedical image analysis mainly focuses on how to extract the knowledge gained while solving problems in source domain and apply it to different but related problems in target domain, thus making the deep models converge quickly, robustly and accurately especially in small sample cases. Although there exists various transfer learning techniques, fine-tuning a deep model that is pre-trained on ImageNet, COCO or PascalVOC might be a good start to perform biomedical image analysis tasks with inadequate annotated samples.

## 5. Active learning techniques

The performance of deep models is gradually improved with the increase of annotated samples before reaching an inflection point, while after that the increase of annotated samples will not improve deep models much (Fig. 7a). Not every annotated sample contributes equally to deep models. Selecting the few most valuable samples to accelerate the arrival of inflection point might be a potential solution to reduce the dependence of deep models on large-scale annotated samples, and that is what active learning techniques expect to do. Due to the high cost of labeling multimodal biomedical images, it is of great value to achieve the performance of deep models trained on small-scale biomedical images comparable to that of deep models trained on large-scale biomedical images. The active learning technique might be a promising approach to achieve this goal.

The problem that active learning techniques expect to solve is how to train an effective model with the least annotation cost. There are generally four key components in active learning framework (Fig. 7b), including low-cost unlabeled samples, an oracle, a task-specific model and a querying strategy, where the oracle plays the role of an annotator (e.g., biomedical expert). At a given point, querying strategy selects a sample that is most valuable to train task-specific model from low-cost unlabeled samples. The selected sample is labeled by oracle and is subsequently added to training set to furtherly train task-specific model. The core problem that has been investigated over years in active learning is how to customize a querying strategy for a specific task. Many querying strategy frameworks have been proposed, involving query by committee, expected model change, uncertainty sampling and expected error reduction, etc. The majority of these querying strategies are designed to query the labels for the most informative sample(Zhou et al., 2017; Zhou et al., 2019; K. Wang et al., 2017; Huang et al., 2014) or (and) the most representative sample(Zhou et al., 2019; K. Wang et al., 2014; S.-J. Huang et al., 2018; Wang et al., 2016). Compared with the random sampling strategy that is generally used to train deep models (also called passive learning), the querying strategy in active learning is able to use the selected and labeled samples to train deep models more efficiently and effectively. The current research of active learning techniques focus on solving complex problems in practical application scenarios, such as active learning with weak supervision(Hua et al., 2018; Xu et al., 2017) , cost-sensitive active learning(S.-J. Huang et al., 2017; Yan and Huang, 2018) and active learning for deep models(Hua et al., 2018; Sener and Savarese, 2017; K. Wang et al., 2017) , etc.

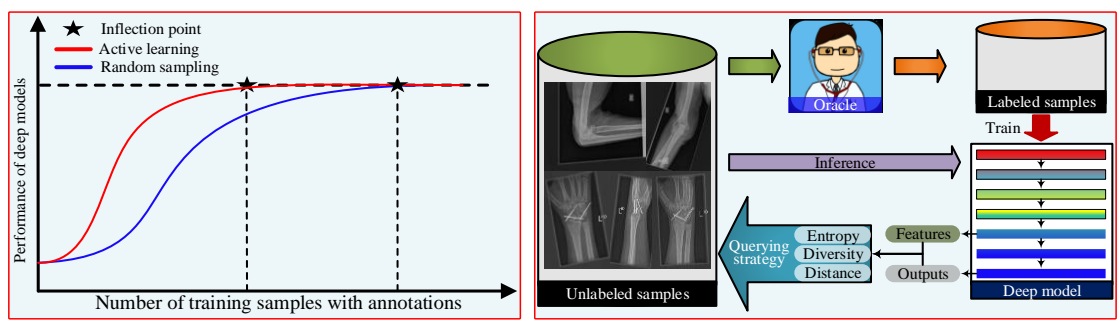

Figure 7. Left-to-right: (a) The learning curve of model performance over training sample size; (b) illustration of active learning framework.

Table 4. Representative papers using active learning techniques in biomedical image analysis based on deep models.

| Reference | Querying strategy | Tasks | Dataset | Deep models |
|---|---|---|---|---|
| Zhou et al. (2017) | Taking the diversity and entropy of model's predictions for augmented unlabeled samples as a sampling criterion | Colonoscopy frame classification | 6 complete colonoscopy videos | AlexNet |
| | | Polyp detection | 38 short colonoscopy videos | |
| | | Pulmonary embolism detection | 121 CT pulmonary angiography datasets with a total of 326 PEs | |
| Yang et al. (2017) | Taking the similarity and uncertainty of unlabeled samples determined by multiple FCNs' predictions as querying strategy for annotation suggestion | Gland segmentation | 165 colon histology images(Sirinukunwattana et al., 2017) | FCN |
| | | Lymph node segmentation | 74 ultrasound images(Zhang et al., 2016b) | |
| Chowdhury et al. (2017) | Uncertainty sampling based on model's confidence in unlabeled patches | Cellular segmentation | NIH-3T3 (mouse embryonic fibroblasts), MCF10A (human breast epithelial cells) and HeLa-S3 (cervix adenocarcinoma)(Van Valen et al., 2016) | Customized CNN-based model |
| Gorriz et al. (2017) | Pixel-wise uncertainty built by using Monte Carlo dropout to deep model during testing on unlabeled samples | Melanoma segmentation | 2000 RGB dermoscopy images(Gutman et al., 2016) | UNet |
| Folmsbee et al. (2018) | Uncertainty sampling based on the qualitative comparison of the model's outputs for unlabeled samples by a trained pathologist | Tissue classification in oral cavity cancer | 143 tissue pathology slides from oral cavity cancer tumor resections | Customized CNN-based model |
| Sourati et al. (2018) | Sampling criterion formulated by diversified Fisher information | Newborn and adolescent brain segmentation | 66 brain MR images of adolescents and 25 brain MRI images of newborns(Makropoulos et al., 2018) | Customized CNN-based model |
| Ozdemir et al. (2018) | Uncertainty sampling based on prediction variance and representativeness sampling based on content distance | Bone and muscle tissue segmentation | 36 MR images of patients diagnosed with rotator cuff tear (shoulders) | DCAN(Chen et al., 2017) |
| Zhou et al. (2018) | Taking the diversity and entropy of model's predictions for augmented unlabeled samples as a sampling criterion | Carotid intima-media thickness video interpretation | 92 carotid intima-media thickness videos from 23 patients | AlexNet |
| Smailagic et al. (2018) | Taking the maximum mean distance between unlabeled and labelled samples in the oriented FAST and rotated BIREF feature space as sampling criterion | Diabetic retinopathy detection | 1200 eye fundus images from 654 diabetic and 546 healthy patients | Inception V3 |
| | | Breast cancer classification | 400 breast tissue cells images(Aresta et al., 2019) | |
| | | Skin cancer classification | 900 benign and malignant cell tissue images(Gutman et al., 2016) | |

In order to dramatically reduce annotation cost, many studies have adopted active learning techniques to train deep models in biomedical image analysis tasks (Table 4), mainly involving classification tasks(Zhou et al., 2017; Zhou et al., 2019; Chowdhury et al., 2017; Folmsbee et al., 2018; Smailagic et al., 2018) and segmentation tasks(Yang et al., 2017; Gorriz et al., 2017; Ozdemir et al., 2018; Sourati et al., 2018). For relatively simple biomedical image classification tasks, the predictions of deep classification models for unlabeled samples are generally useful for active learning techniques. Many querying strategies have formulated the criterion for uncertainty sampling based on model's predictions, such as the symmetric Kullback Leibler divergence and entropy of model's predictions for augmented samples(Zhou et al., 2017; Zhou et al., 2019), model's confidence in unlabeled patches(Chowdhury et al., 2017) and subjective comparison of CNN's predictions from an experienced pathologist(Folmsbee et al., 2018). Regarding to relatively complex biomedical image segmentation tasks, the uncertainty of model's prediction mask for individual unlabeled sample and the representativeness amongst unlabeled samples based on the distance of deep features are generally combined to formulate the sampling criterion for querying strategy(Yang et al., 2017; Ozdemir et al., 2018; Sourati et al., 2018). To formulate a more effective criterion for uncertainty sampling, there generally exists three major approaches, where multiple predictions are required for the same candidate unlabeled sample. First, under the consideration that an uncertain sample is prone to making deep models sensitive to data augmentation, the sampling criterion can be formulated based on model's predictions for multiple augmented samples of a candidate(Zhou et al., 2017; Zhou et al., 2019). Second, with the claim that an uncertain sample is prone to being sensitive to dropout operation applied to CNN while a confident sample is generally less sensitive to dropout, the sampling criterion can be formulated based on the inconsistency of model's predictions at different random dropout conditions(Gorriz et al., 2017; Ozdemir et al., 2018). Third, the strategy of query by committee trains multiple deep models to construct a committee and thus adopt the variance of these models' predictions as sampling criterion(Yang et al., 2017).

Existing studies using active learning techniques to train deep models on multimodal biomedical images have achieved impressive results in reducing the required annotation cost. This key technique for deep learning of small sample data mainly focuses on how to customize a querying strategy for a specific biomedical image analysis task that can select the few most valuable samples for annotation, and how to effectively train deep models with these actively selected samples. However, querying for the few most valuable samples from unlabeled samples is computationally expensive in general and in practice, training deep models with selected samples effectively is very challenging. Combining active learning techniques and transfer learning techniques(Zhou et al., 2017; Zhou et al., 2019) might be one promising research method. Such combination has the potential to make computer-aided diagnosis system based on deep models incrementally evolve through online feedback of false instances corrected by biomedical experts, where the challenge is how to deal with catastrophic forgetting of false instances.

## 6. Miscellaneous techniques

### 6.1 Domain knowledge

Biomedical image analysis is a highly specialized work, which requires large expertise. In general, there is a large amount of prior domain knowledge in a specific biomedical image analysis task. The prior domain knowledge is able to regularize deep models in a certain sense, which can narrow the parameter space of deep models and thus relieve overfitting. Therefore, combining domain knowledge might be a promising approach to alleviate the dependence of deep learning on large-scale labeled samples in medical image analysis applications at the meantime of improving the robustness of deep models.

Table 5. Representative papers using domain knowledge in biomedical image analysis based on deep models.

| Reference | Knowledge & method | Tasks |
|---|---|---|
| Xie et al. (2018)(2019) | The malignant nodule generally has a lobed contour, an unsmooth edge and an inhomogeneous density; characterized and digitized by hand-crafted features, and furthery fused with deep features in decision-level. | Benign-malignant nodule diagnosis in pulmonary CT images |
| Yan et al. (2015) | Radiologists usually recognize the body part of a slice based on local discriminative regions rather than global image context; used as a design guide for multi-stage deep network. | CT slice-based body part recognition |
| Fu et al. (2019) | Global anterior segment structure, local iris region and anterior chamber angle (ACA) patch are clinically important to angle-closure detection; used as a design guide for multilevel deep network. | Closed-angle detection in anterior segment OCT (AS-OCT) images |

There are two main approaches to combine domain knowledge and deep learning in biomedical image analysis tasks (Table 5): (1) the domain knowledge is characterized and digitized by hand-crafted features, which is furthery fused with deep features(Xie et al., 2019; Xie et al., 2018) , and (2) the domain knowledge is used as a design guide for deep networks(Yan et al., 2015; Fu et al., 2019). Deep models using prior domain knowledge in these studies(Xie et al., 2019; Xie et al., 2018; Yan et al., 2015; Fu et al., 2019) consistently achieved superior performance over standard methods.

Domain knowledge is important to train deep models for a specific biomedical image analysis task. However, the generality and end-to-end learning framework of deep models suffers due to the use of task-specific domain knowledge. Besides, distilling prior domain knowledge requires years of clinical experience in biomedical image analysis, where only experienced biomedical experts can do that. Thus, digging deep into clinical expertise underlying biomedical images through collaboration with clinicians might be a good start to apply deep learning techniques in biomedical image analysis tasks with inadequate annotated samples.

## 6.2 Shallow methods

Compared with the traditional shallow methods based on hand-crafted features, deep learning techniques have made significant progress in various computer vision applications, which relies heavily on large-scale labeled samples. With inadequate training samples, the deep models are prone to suffering from overfitting, thus leading to a big drop in performance. In comparison of deep learning techniques, the shallow methods generally have lower dependency on large-scale labeled samples due to the more general hand-crafted features and lower model complexity. A natural research idea is to combine shallow methods and deep models to alleviate the dependence on large-scale labeled samples and improve performance.

Shallow methods and deep models can be combined in four different levels for biomedical image analysis (Table 6), including input-level(Wang et al., 2014) (e.g., original image and edge image), feature-level(Lai and Deng, 2018) (e.g., deep features and hand-crafted features), decision-level(Xie et al., 2018; Wang et al., 2014; Jianpeng Zhang et al., 2018) (e.g., CNN-based classifier and SVM) and mixed-level(Wang et al., 2014) (e.g., SVM classifier based on deep features). The combined model is expected to inherit the advantages of both shallow methods and deep learning techniques to improve the analysis performance and generalization especially in small sample cases.

Table 6. Representative papers combining shallow methods and deep models in biomedical image analysis.

| Reference | Fusion level & method | Tasks |
|---|---|---|
| Wang et al. (2014) | Input-level; edges based on a thresholding method and Laplacian of Gaussian + CNN-based classifier | Mitosis detection in breast cancer pathology images |
| | Mixed-level; random forest classifier + the deep features | |
| | Decision-level; 2 random forest classifiers + a CNN-based classifier | |
| Xie et al. (2018) | Decision-level; texture descriptor based on gray level co-occurrence matrix (GLCM), Fourier shape descriptor and CNN-based deep features are fused by three AdaBoosted backpropagation neural networks (BPNN) | Benign-malignant nodule diagnosis in pulmonary CT images |
| Zhang et al. (2018) | Decision-level; the bag of feature (BoF), local binary pattern (LBP) features and CNN-based deep features are fused by BPNN | ImageCLEF 2016 medical classification task(García Seco de Herrera et al., 2016) |
| Lai et al. (2018) | Feature-level; hand-crafted features based on texture and color statistics and CNN-based deep features are fused by the multilayer perceptron | Skin lesion classification and tissue classification tasks |

In summary, combining traditional shallow methods and deep models is a promising approach to apply deep learning of small sample in biomedical image analysis. This key technique mainly focuses on how to select appropriate hand-crafted features and design the fusion strategy for a specific biomedical image analysis task. However, a key limitation of these studies is that the fusion strategy is so tricky that an inappropriate combination might aggravate overfitting problem in small sample cases, thus leading to a big drop in performance. Prior domain knowledge might be a good design guide for hand-crafted features and fusion strategy.

## 6.3 Attention mechanism

Table 7. Representative papers using attention mechanism in biomedical image analysis based on deep models.

| Reference | Method | Tasks |
|---|---|---|
| Wang et al. (2018) | Adding an attention network branch based on Grad-CAM to a ResNet-152 network | Lung disease diagnosis in chest X-ray images |
| Schlemper et al. (2018) | Adding two designed attention-gated branches to a CNN-based network | Ultrasound scan plane detection |
| Schlemper et al. (2018) | Adding attention-gated skip connections to a UNet | Pancreas segmentation in abdominal 3D CT scans |
| Zhang et al. (2019) | Using designed attention residual learning (ARL) block | Skin lesion classification |

Deep models show an attention mechanism of background suppression and foreground enhancement in feedforward operation. As mentioned earlier, this attention mechanism can be used for visual explanation and weakly supervised object detection and segmentation. The attention mechanism is a useful tool to refine the deep feature representations. Through regulating the attention of deep models, deep models can observe input image more "carefully", which is expected to improve the performance of deep learning in small sample cases.

In some biomedical image analysis tasks (Table 7), researchers have tried to regulate the attention of deep models by adding attention branch network(Wang and Xia, 2018; Schlemper et al., 2018; Oktay et al., 2018) or self-attention blocks(Zhang et al., 2019) to standard deep models. These methods notably improve the analysis performance in comparison of counterparts without attention regulation. Properly regulating the attention of the deep models can be very helpful to biomedical image analysis. The attention mechanism enables deep models to focus on both global and local representation of input images simultaneously without using any extra localization modules. Such ability is equivalent to stimulating deep models to observe the input image more "carefully", which is much more valuable for biomedical image analysis in small sample cases. Besides, this technique might be an available approach to extend weakly supervised learning techniques (e.g., weakly supervised object localization) by utilizing the (intermediate) results of weakly supervised tasks (e.g., attention map or lesion location) to improve the performance of original tasks (e.g., image classification).

## 6.4 Data Augmentation

Visual appearance of the object of interest generally experiences various variations, involving illumination variation, scale variation, ration, translation and deformation, etc. Machine learning theory claims that training a model on an image set with higher diversity is prone to getting a more robust and generalized model. However, collecting and labeling all the images that involve these variations seems to be an impossible task. Data augmentation techniques are thus proposed to increase both the amount and diversity of training samples by applying various transformations (e.g., horizontally flipping) to original samples. Intuitively, data augmentation is used to teach a model about invariances behind the variations in data domain(Cubuk et al., 2019). Many data augmentation techniques have been proposed with the development of deep learning, concisely categorized as classical transformations (Rajpurkar et al., 2018; Rajpurkar et al., 2017) , generative adversarial network (GAN)(S.-W. Huang et al., 2018; Pan et al., 2018) and learning to augment(Cubuk et al., 2019; Perez and Wang, 2017).

**Classical transformations**. Classical transformations mainly involve horizontal flipping, random cropping (extracting patches), random scaling, random rotating, translations, shearing and elastic deformations, which can be performed on-the-fly or off-line. In general, the classical transformations are widely adopted in deep learning techniques and the best augmentation strategies are dataset-specific and task-specific. To be more concise, we survey the classical transformation techniques based on some representative applications in biomedical image classification and segmentation tasks (Table 8). Random flipping, random rotating, and random cropping are the three most frequently-used classical transformations. Moreover, extracting patches from original images is prone to being used to augment training samples from high-resolution biomedical images, especially in small sample cases. Besides, elastic deformations seem to be important to biomedical image segmentation tasks since deformations are the most common variations in tissue as Ronneberger et al.(2015) pointed out.

**GAN**. GAN techniques have been widely used to synthesize biomedical images(Pan et al., 2018; Huo et al., 2018; Chen et al., 2019). The synthetic biomedical images show a remarkable competitiveness in data augmentations due to their variety and reality in appearance variations. Biomedical images are generally synthesized from one modality to another, such as synthesizing PET from MRI in diagnosis of Alzheimer's disease (AD)(Pan et al., 2018) and synthesizing CT from MR in cross-modality medical image segmentation tasks(Huo et al., 2018; Chen et al., 2019). The synthetic images can be furtherly used for all kinds of biomedical image analysis with or without images from source modality. For further exploration, various applications of GANs in biomedical imaging can be found in Yi's work(Yi et al., 2018). Although the GAN techniques have achieved many successes in biomedical image synthesis, it is worth noting that there still exists lots of risks involving non-convergence, mode collapse and highly sensitive hyper-parameter selections.

**Learning to augment**. Compared with classical transformations, learning to augment aims to find the most effective data augmentation strategy automatically, which is sort of similar to the active sampling strategy in active learning techniques. For convenience, the majority of existing studies(Cubuk et al., 2019; Perez and Wang, 2017) evaluate their methods of learning to augment on natural datasets. Inspired by these studies, Zhao et al.(2019) proposed an automated data augmentation method for one-shot brain MRI segmentation. Specifically, they designed two independent spatial and appearance transformation models based on the UNet architecture(Ronneberger et al., 2015) to learn how to implement natural variations in MRI scans. The two transformation models were sequentially used to generate realistic labeled MRI scans to train segmentation model. This method achieved superior performance over state-of-the-art methods for one-shot biomedical image segmentation.

Data augmentation is very important to deep learning techniques that are applied in biomedical image analysis, especially in small sample cases. The augmentation techniques are able to increase both the amount and diversity of training samples with lower cost in comparison of manual acquisition. Proper augmentation helps relieve the suffering of deep models from small sample learning and make deep models robust to variations. However, aggressive and improper augmentation might break the nature distribution of training samples, thus leading to a notable drop in the analysis performance of practical applications. Therefore, a soft and stepwise augmentation strategy is suggested to be used in biomedical image analysis.

Table 8. Classical transformations applied in some representative biomedical image classification and segmentation tasks based on deep models.

| Task | Reference | Dataset | Data augmentation strategy |
|---|---|---|---|
| Classification | Anthimopoulos et al. (2016) | 135 CT scans with 512×512 pixels per slice | Extracting patches, random flipping and random rotating |
| | Tajbakhsh et al. (2016) | 40 short colonoscopy videos | Extracting patches, scaling, translation, horizontal and vertical mirroring and flipping |
| | | 121 CT pulmonary angiography datasets with a total of 326 PEs | Extracting patches, translation and rotating |
| | | 6 complete colonoscopy videos | Extracting patches |
| | Esteva et al. (2017) | 129,450 skin lesion images comprising 2,032 different diseases | Random rotating, cropping and vertical flipping |
| | Gondal et al. (2017) | 88,702 fundus images in Kaggle Dataset and 89 fundus images in DiaretDB1 Dataset(Kauppi et al., 2007, p. 1). | Rotating, horizontal and vertical flipping |
| | Quellec et al. (2017) | 88,702 fundus images in Kaggle Dataset and 89 fundus images in DiaretDB1 Dataset(Kauppi et al., 2007, p. 1). | Rotating, translation, scaling, horizontal flipping and cropping |
| | Zhou et al. (2017) | 6 complete colonoscopy videos | Extracting patches |
| | | 38 short colonoscopy videos | Scaling, translation, mirroring and flipping |
| | | 121 CT pulmonary angiography datasets with a total of 326 PEs | Extracting patches, scaling, translation and rotating |
| | Samala et al. (2017) | 1,655 SFM views and 310 DM views with 2454 masses (1057 malignant, 1397 benign) | Flipping and rotating |
| | Rajpurkar et al. (2018) | 40,561 multi-view musculoskeletal radiographs | Random horizontal flipping and rotating |
| | Fernando et al. (2018) | 1300 skin lesion images(Ballerini et al., 2013) | Web-crawling |
| | Smailagic et al. (2018) | 1200 eye fundus images from 654 diabetic and 546 healthy patients | Cropping, flipping and translation |
| | | 400 breast tissue cells images(Aresta et al., 2019) | |
| | | 900 benign and malignant cell tissue images(Gutman et al., 2016) | |
| | Xie et al. (2018) | 1018 clinical chest CT scans with lung nodules(Armato et al., 2011) | Random translation, rotating and horizontal or vertical flipping |
| | Lai et al. (2018) | 2828 fundamental tissue images in HIS2828 dataset | Randomly cropping, horizontal or vertical flipping |
| | | 2000 skin lesion images in ISIC2017 dataset(Codella et al., 2018) | |
| | Schlemper et al. (2018) | 2694 2D ultrasound examinations of volunteers with gestational ages between 18 and 22 weeks | Horizontal and vertical translation, horizontal flipping, rotating and scaling |
| | Zhang et al. (2019) | 2750 skin dermoscopy images in ISIC2017 dataset(Codella et al., 2018) | Random rotation, scaling, horizontal and vertical flip |
| Segmentation | Ronneberger et al. (2015) | 30 images (512x512 pixels) from serial section transmission electron microscopy | Elastic deformations and drop-out |
| | | 35 partially annotated light microscopic images | |
| | | 20 partially annotated light microscopic images | |
| | Chen et al. (2016) | 85 histopathological images (benign / malignant = 37 / 48) | Translation, rotating, and elastic deformations |
| | Kamnitsas et al. (2016) | 61 brain MRI sequences | Adding images reflected along the sagittal axis and intensity shifting |
| | Isensee et al. (2017) | 135 MRI scans of glioblastoma and 108 MRI scans of lower grade glioma | Random rotating, scaling, elastic deformations, gamma correction augmentation and mirroring |
| | Schlemper et al. (2018) | 150 gastric cancer 3D CT scans(Roth et al., 2017) | Affine transformations, horizontal flipping and random cropping |
| | | 82 contrast enhanced 3D CT scans with pancreas(Roth et al., 2016) | |
| | Laves et al. (2018) | 536 laryngeal endoscopic images | Horizontal flipping and rotating |
| | Kervadec et al. (2019) | Prostate transversal T2-weighted MR images of 50 patients (https://promise12.grand-challenge.org) | Mirroring, flipping and rotating |
| | Isensee et al. (2019) | 100 cine-MRI images | Elastic deformations, random scaling, random rotating, gamma augmentation, and spatial transformations |
| | | 131 CT images | |
| | | 30 abdominal CT images | |
| | | 50 anisotropic MRI images | |

## 7. Conclusion

A comprehensive survey on key SSL techniques for deep learning of small sample in clinical biomedical image analysis is presented. First, we introduce explanation techniques for deep models that are expected to satisfy the demands of clinical diagnosis for explainable and transparent decision-making process. Second, we present weakly supervised learning techniques to deal with the problem how to effectively train deep models with coarse-grained annotations for fine-grained biomedical image analysis tasks. Third, transfer learning techniques are surveyed to answer this question how to retrieve the knowledge gained by deep models while solving problems in other domains and reuse it in biomedical image analysis. Fourth, active learning techniques are discussed to solve the problem how to select the few most valuable samples for annotations and train deep models with them for optimal performance in biomedical image analysis. Finally, we study miscellaneous techniques that are also important to apply deep models for biomedical image analysis, involving data augmentation, domain knowledge, traditional shallow methods and attention mechanism. These key techniques are expected to effectively support the application of deep learning in clinical biomedical image analysis, and furthery improve the analysis performance, especially when large-scale annotated samples are not available. To our best knowledge, this is the first attempt to present a comprehensive survey on key SSL techniques for deep learning of small sample in biomedical image analysis through combining with the development of related techniques in computer vision applications. Notably, our survey focusing on key SSL techniques can be mutually complementary with previous reviews on deep learning techniques in biomedical image analysis.

## Acknowledgements

The authors would like to thank members of the Medical Image Analysis for discussions and suggestions. This research did not receive any specific grant from funding agencies in the public, commercial, or not-for-profit sectors.